\pdfoutput=1
\documentclass[10pt,letterpaper]{article}

\usepackage{epigenetics-preprint}
\definecolor{newtext}{HTML}{C0392B}

\preprintKicker{bioRxiv preprint}
\preprintCategory{Computational Biology}
\preprintDate{June 2026}
\preprintTitle{EpiBench: Verifiable Evaluation of AI Agents on Epigenomics Analysis}
\preprintAuthors{Harihara Muralidharan\preprintSup{1}, Reema Baskar\preprintSup{1}, Soo Hee Lee\preprintSup{1}, Tim Proctor\preprintSup{1}, Kenny Workman\preprintSup{1}}
\preprintAffiliations{\preprintSup{1}LatchBio, San Francisco, CA, USA}
\preprintCorrespondence{kenny@latch.bio}

\begin{document}

\PreprintHeader

\begin{PreprintAbstract}
We introduce EpiBench, a verifiable benchmark for short-horizon epigenomics
analysis. EpiBench evaluates whether agents can make well-defined
analysis decisions from realistic workflow states and return deterministically
gradable answers. The benchmark includes 106 evaluations across
CUT\&Tag/CUT\&RUN, ATAC-seq, ChIP-seq, and DNA methylation workflows. Across
5,088 valid trajectories from 16 model-harness pairs, no system passed a
majority of attempts: GPT-5.5 / Pi led at 45.0\% (143/318 attempts; 95\%
confidence interval (CI), 36.3--53.7), followed by GPT-5.5 / OpenAI Codex at
39.9\% (127/318 attempts; 95\% CI, 31.6--48.3). Claude Opus 4.8 Max / Pi and
GPT-5.4 / Pi each passed 39.0\% (124/318 attempts; 95\% CI, 30.2--47.8 and
31.0--47.0, respectively). Performance varies across assay
types, and many failed runs still contain parts of the correct answer. Agents
often found the right files and computed useful intermediate results, but
failed when the task required deeper, assay-specific scientific judgment.
\end{PreprintAbstract}

\vfill
\clearpage

\section*{Topline Benchmark Performance}

{\fontsize{9.1}{12.8}\selectfont
We ran the benchmark across 16 model-harness pairs. Pi denotes the Pi terminal
coding harness. Passing requires exact recovery of the graded structured
answer for an evaluation attempt. The result set contains 5,088 valid
trajectories: each of the 16 model-harness pairs was run three times on each
of the 106 evaluations, yielding 318 attempts per pair.
\par}

\begin{center}
  \includegraphics[width=0.94\textwidth]{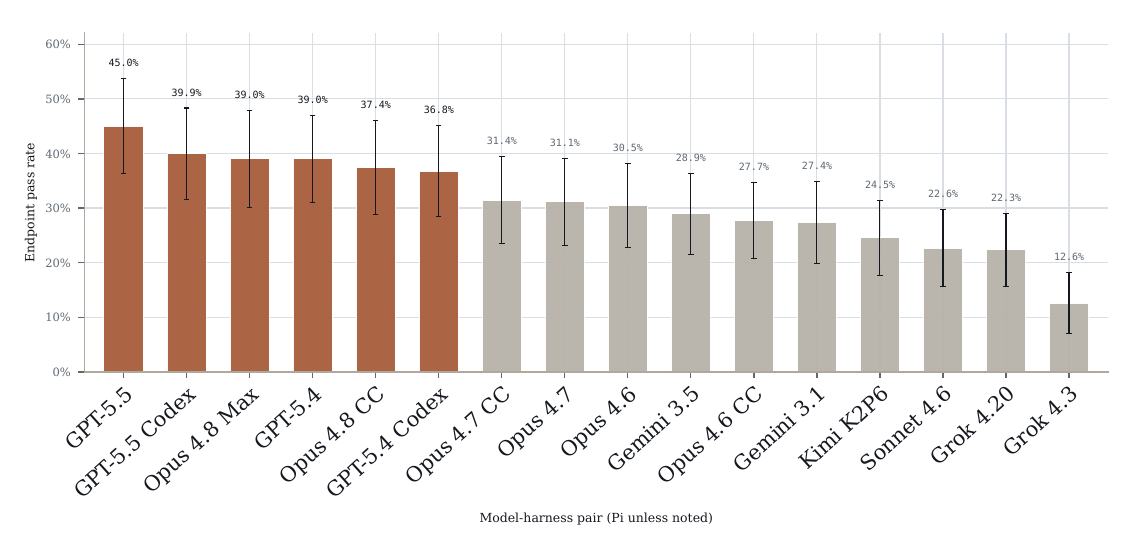}

  \vspace{0.18in}

  \begin{minipage}[t]{0.62\textwidth}
    \centering
    \includegraphics[width=\linewidth]{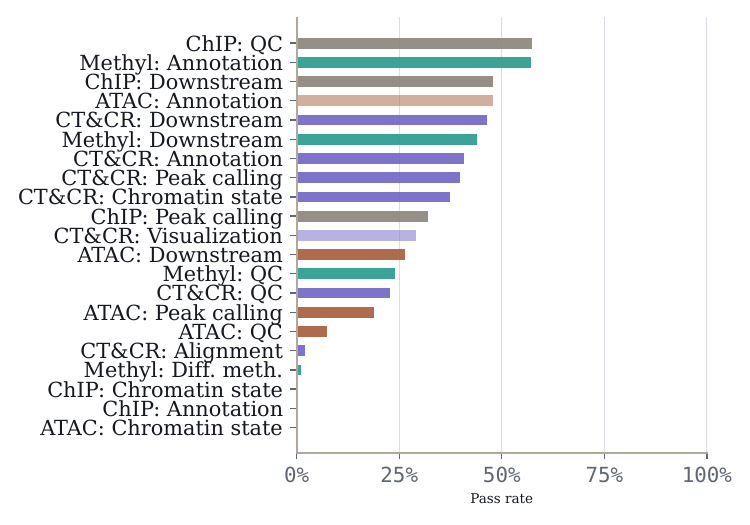}
  \end{minipage}
  \hfill
  \begin{minipage}[t]{0.36\textwidth}
    \centering
    \includegraphics[width=\linewidth]{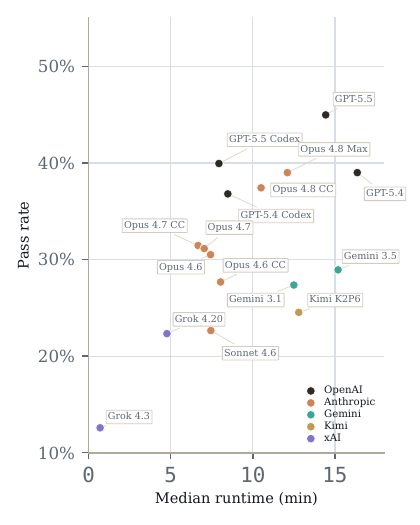}
  \end{minipage}

  \captionof{figure}{\textbf{Topline benchmark performance.}
  Top: evaluation-weighted pass rate by model-harness pair across all trajectories.
  The top panel shows Student $t$ confidence intervals over evaluation-level
  mean pass rates. Bottom left: pass rate by assay and task family. Bottom
  right: pass rate against median runtime. Evaluation-level attempt
  statistics are reported in Table~\ref{tab:model-results}.}
  \label{fig:topline-performance}
\end{center}

\clearpage

{\fontsize{9.1}{12.8}\selectfont
\newcommand{\TwoColParaBreak}{\par\vspace{0.62em}\noindent}
\noindent\begin{minipage}[t]{0.47\textwidth}
\setlength{\parskip}{0pt}
\setlength{\parindent}{0pt}

\section{Introduction}

Epigenomic assays measure the regulatory state of the genome: chromatin
accessibility, histone modifications, DNA methylation, and other molecular
signals that link genotype to cell state and gene regulation
\cite{encode2012,roadmap2015}. But raw epigenomics data are not directly
interpretable. Drawing scientific conclusions requires a sequence of analysis
decisions across read alignment, peak calling, motif interpretation, sample
aggregation, normalization, and genomic annotation. Small procedural choices
can change the final biological conclusion.

\TwoColParaBreak
AI agents show early promise in human-guided analysis of biological data. They
can inspect files, write code, invoke command-line tools, and exercise some
scientific judgment in analysis tasks
\cite{bixbench2025,compbiobench2026,biomysterybench2026}. But they remain prone to hallucination
and unreliable biological judgment, especially across diverse tissues,
diseases, study designs, and assay types.

\TwoColParaBreak
Existing biology benchmarks increasingly test practical analysis in other data
modalities or at broader biological scope
\cite{labbench2024,bixbench2025,biomnibench2026,compbiobench2026,genebench2026,workman2025spatialbench,workman2026scbench},
but, to our
knowledge, none focus on epigenomics analysis alone. We introduce EpiBench, a
verifiable benchmark for short-horizon epigenomics tasks. Each evaluation
snapshots a realistic workflow state immediately before a target result. The
agent receives relevant files, metadata, and task context, then must interact
with the data and return a structured answer graded deterministically. The
benchmark includes 106 evaluations across CUT\&Tag/CUT\&RUN, ATAC-seq,
ChIP-seq, and DNA methylation workflows, covering quality control, peak
calling, chromatin-state analysis, genomic annotation, differential
methylation, downstream quantitative analysis, alignment, and visualization.

\TwoColParaBreak
Across 5,088 valid trajectories from 16 model-harness pairs, no system passed
a majority of attempts. GPT-5.5 / Pi led at 45.0\% (143/318 attempts; 95\%
CI, 36.3--53.7), followed by GPT-5.5 / OpenAI Codex at 39.9\% (127/318
attempts; 95\% CI, 31.6--48.3). Claude Opus 4.8 Max / Pi and GPT-5.4 / Pi
each passed 39.0\% (124/318 attempts; 95\% CI, 30.2--47.8 and 31.0--47.0,
respectively). The failures show that current agents can sometimes make reasonable
progress with
epigenomics data, but still struggle with some analysis decisions, especially
those that require biological
reasoning: choosing which samples to compare, accurate normalization,
identifying appropriate genomic features, and reporting appropriate
statistics.

\end{minipage}\hfill
\begin{minipage}[t]{0.47\textwidth}
\setlength{\parskip}{0pt}
\setlength{\parindent}{0pt}

\section{Benchmark Construction}

To construct a benchmark that approximates real epigenomics analysis, we
derived evaluations from practical workflows across CUT\&Tag/CUT\&RUN,
ATAC-seq, ChIP-seq, and DNA methylation. We decomposed analysis workflows
into short, gradeable decisions: quality control, peak calling,
chromatin-state comparisons, genomic annotations, methylation statistics,
alignment checks, and downstream quantitative analyses such as motif,
gene-set, regulatory-domain, and cross-modality summaries. At each step, we
sought to isolate the empirical decision that determines the result and
formalize it as a deterministically gradable answer.

\TwoColParaBreak
The final evaluation suite consists of 106 problems spanning four assay
families and eight task categories. Each problem includes a snapshot of real
assay outputs immediately before a target analysis step, metadata needed to
interpret the files, a high-level task description, and a deterministic grader
that evaluates whether the agent recovered the requested empirical result.

\TwoColParaBreak
We follow three design criteria. First, tasks must be verifiable: each
evaluation admits an automatically checkable success condition, such as a
numerical interval check, structured label match, or all-of field comparison.
Second, tasks should be scientifically durable: the intended answer should be
recoverable across reasonable analysis choices, so the benchmark measures
epigenomics reasoning rather than incidental implementation details. Third,
tasks should resist shortcuts: agents must interact with the provided data
artifacts, and we remove items that can be solved reliably by trivial
heuristics or prior biological knowledge alone.

\TwoColParaBreak
Tasks are designed to specify what should be recovered rather than how it
should be recovered. Some procedural evaluations name a specific operation,
but most scientific evaluations leave the agent to choose reasonable analysis
methods to accomplish a goal. A correct answer should therefore reflect the
result supported by the specific assay context, not a nearby generic workflow
convention or a plausible biological expectation.

\TwoColParaBreak
All evaluations underwent manual quality control. We inspected agent
trajectories across runs to identify prompts that over-specified the method,
answers that could be obtained through shortcuts, and graders that failed to
distinguish supported results from plausible but unsupported biological
outputs.
\par

\end{minipage}
}

\clearpage

{\fontsize{9.1}{12.8}\selectfont
\section{Evaluation Inventory}

\begin{multicols}{2}
The benchmark covers four assay types and eight task categories.
CUT\&Tag/CUT\&RUN tasks are drawn from zebrafish chromatin workflow snapshots
and span chromatin state analysis, QC, peak calling, downstream analysis,
genomic annotation, alignment, and visualization. ATAC-seq tasks use pediatric
B-ALL ATAC/RNA data from GSE161501 and related integrative workflow
artifacts \cite{diedrich2021allatac}, with downstream analysis dominating.
ChIP-seq tasks use B-ALL H3K27ac data from GSE211631
\cite{barnett2023ball} and include peak calling, chromatin state analysis, QC,
genomic annotation, and downstream analysis. Methylation tasks are derived
from the ESCC WGBS/RNA data GSE149608 and GSE149609 \cite{cao2020escc},
covering QC, downstream analysis, genomic annotation, and differential
methylation.
\par

{\fontsize{8.3}{11.4}\selectfont
Across all 106 evaluations, 34 are downstream analysis tasks, 22 are QC, 16
are chromatin state analysis, 14 are peak calling, 12 are genomic annotation,
4 are differential methylation, 3 are alignment, and 1 is visualization. The
result set contains 5,088 valid trajectories from 16 model-harness pairs,
with three attempts for each evaluation and model-harness pair.
\par}

\end{multicols}
}

\begin{center}
\fontsize{7.2}{8.7}\selectfont
\setlength{\tabcolsep}{3.2pt}
\begin{tabular}{@{}>{\raggedright\arraybackslash}p{0.16\linewidth}
                >{\centering\arraybackslash}p{0.055\linewidth}
                >{\raggedright\arraybackslash}p{0.25\linewidth}
                >{\raggedright\arraybackslash}p{0.42\linewidth}@{}}
\toprule
Assay type & Evals & Source data & Task categories \\
\midrule
CUT\&Tag/CUT\&RUN &
47 &
Zebrafish chromatin workflow snapshots &
Chromatin state (12), QC (10), peak calling (9), downstream analysis (7), genomic annotation (5), alignment (3), visualization (1) \\
\addlinespace
ATAC-seq &
24 &
B-ALL ATAC/RNA (GSE161501 and integrative workflow artifacts) &
Downstream analysis (17), chromatin state (2), peak calling (2), QC (2), genomic annotation (1) \\
\addlinespace
ChIP-seq &
10 &
B-ALL H3K27ac (GSE211631) &
Peak calling (3), chromatin state (2), QC (2), downstream analysis (2), genomic annotation (1) \\
\addlinespace
Methylation-seq &
25 &
GSE149608, GSE149609 &
QC (8), downstream analysis (8), genomic annotation (5), differential methylation (4) \\
\bottomrule
\end{tabular}
\captionof{table}{\textbf{Evaluation inventory.}
The table lists the number of evaluations, source data, and task categories
for each assay type.}
\label{tab:evaluation-inventory}
\end{center}

\clearpage

\StartBody

\section{Results}

\subsection{No agent reaches a 50\% endpoint pass rate}

Each of the 16 model-harness pairs was evaluated on the same 106 evaluations
with three attempts per evaluation. No model-harness pair reached a 50\%
endpoint pass rate. GPT-5.5 / Pi had the highest observed pass rate, passing
45.0\% of endpoint attempts (143/318 attempts; 95\% confidence interval (CI),
36.3--53.7), followed by GPT-5.5 / OpenAI Codex at 39.9\% (127/318 attempts;
95\% CI, 31.6--48.3). Claude Opus 4.8 Max / Pi and GPT-5.4 / Pi each passed
39.0\% (124/318 attempts; 95\% CI, 30.2--47.8 and 31.0--47.0, respectively).

The leading systems form a broad tier rather than a stable ordering: their
confidence intervals overlap, and their pass rates are close. Even the strongest
model-harness pair passed fewer than half of endpoint-graded attempts, indicating
that short-horizon epigenomics analysis remains difficult for current agents.

\EndBody

\begin{center}
  \includegraphics[width=0.94\textwidth]{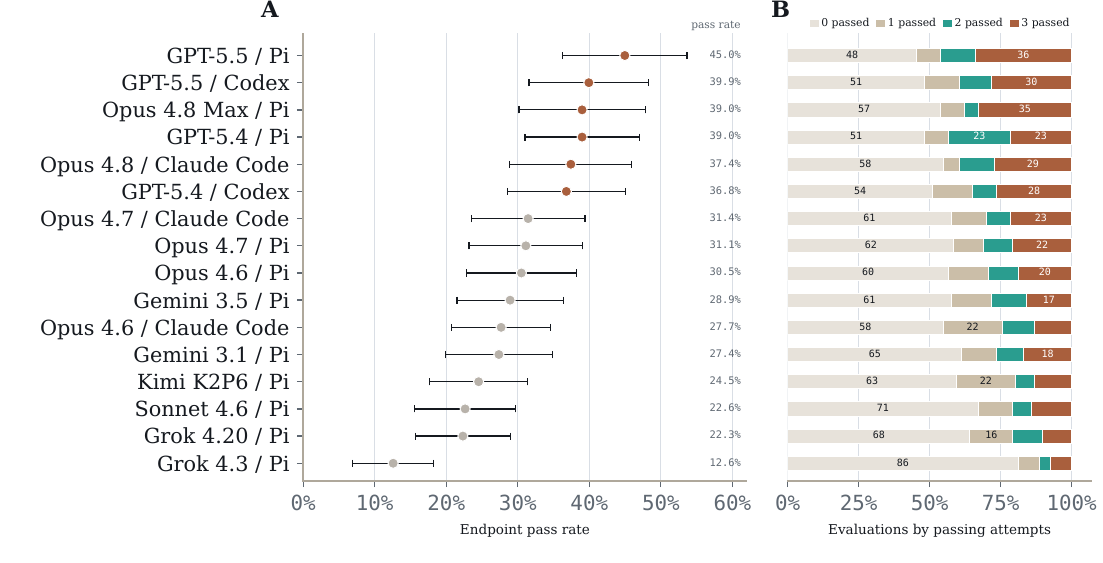}

  \captionof{figure}{\textbf{Endpoint pass rates across model-harness pairs.}
  Panel A: endpoint pass rate across all 5,088 trajectories with 95\%
  confidence intervals. Panel B: evaluation-level attempt outcomes over the same
  model order, partitioning the 106 evaluations for each model-harness pair by
  whether zero, one, two, or all three attempts passed.}
  \label{fig:main-results-summary}
\end{center}

\begin{center}
{\fontsize{5.1}{6.3}\selectfont
\begin{tabular}{@{}>{\raggedright\arraybackslash}p{0.36\linewidth}
                >{\raggedright\arraybackslash}p{0.27\linewidth}rrr@{}}
\toprule
Model / harness & Endpoint pass rate & $\geq$1/3 & $\geq$2/3 & 3/3 \\
\midrule
GPT-5.5 / Pi & 45.0\% (143/318; 36.3--53.7) & 58/106 & 49/106 & 36/106 \\
GPT-5.5 / OpenAI Codex & 39.9\% (127/318; 31.6--48.3) & 55/106 & 42/106 & 30/106 \\
Claude Opus 4.8 Max / Pi & 39.0\% (124/318; 30.2--47.8) & 49/106 & 40/106 & 35/106 \\
GPT-5.4 / Pi & 39.0\% (124/318; 31.0--47.0) & 55/106 & 46/106 & 23/106 \\
Claude Opus 4.8 / Claude Code & 37.4\% (119/318; 28.9--46.0) & 48/106 & 42/106 & 29/106 \\
GPT-5.4 / OpenAI Codex & 36.8\% (117/318; 28.5--45.1) & 52/106 & 37/106 & 28/106 \\
Claude Opus 4.7 / Claude Code & 31.4\% (100/318; 23.5--39.4) & 45/106 & 32/106 & 23/106 \\
Claude Opus 4.7 / Pi & 31.1\% (99/318; 23.2--39.1) & 44/106 & 33/106 & 22/106 \\
Claude Opus 4.6 / Pi & 30.5\% (97/318; 22.8--38.2) & 46/106 & 31/106 & 20/106 \\
Gemini 3.5 Flash / Pi & 28.9\% (92/318; 21.5--36.3) & 45/106 & 30/106 & 17/106 \\
Claude Opus 4.6 / Claude Code & 27.7\% (88/318; 20.7--34.6) & 48/106 & 26/106 & 14/106 \\
Gemini 3.1 Pro / Pi & 27.4\% (87/318; 19.9--34.9) & 41/106 & 28/106 & 18/106 \\
Kimi K2P6 / Pi & 24.5\% (78/318; 17.7--31.4) & 43/106 & 21/106 & 14/106 \\
Claude Sonnet 4.6 / Pi & 22.6\% (72/318; 15.6--29.7) & 35/106 & 22/106 & 15/106 \\
Grok 4.20 reasoning / Pi & 22.3\% (71/318; 15.7--28.9) & 38/106 & 22/106 & 11/106 \\
Grok 4.3 / Pi & 12.6\% (40/318; 6.9--18.2) & 20/106 & 12/106 & 8/106 \\
\bottomrule
\end{tabular}
}

\captionof{table}{\textbf{Endpoint pass rates by model-harness pair.}
Endpoint pass rates are reported as percentage (passing attempts/318 attempts;
95\% CI). The 95\% CI is computed over evaluation-level mean pass rates. The last three columns
report the number of evaluations, out of 106, with at least one, at least two,
or all three attempts passing. Runtime and cost are treated as
operational diagnostics rather than leaderboard columns.}
\label{tab:model-results}
\end{center}

\StartBody

\subsection{Endpoint pass rates vary by assay type}

CUT\&Tag/CUT\&RUN had the highest aggregate pass rate at 34.0\% (768/2,256
attempts; 95\% CI, 25.9--42.2), followed by methylation-seq at 33.3\%
(400/1,200 attempts; 95\% CI, 19.7--47.0) and ChIP-seq at 30.6\% (147/480
attempts; 95\% CI, 9.1--52.2). ATAC-seq was lowest at 22.8\% (263/1,152
attempts; 95\% CI, 10.4--35.2).
Thus, the largest separation was between ATAC-seq and the other assay
families, rather than between a single high-performing assay and the rest of
the benchmark.

The assay groups differed in both size and task composition. CUT\&Tag/CUT\&RUN
contributed 47 evaluations, methylation-seq 25, ATAC-seq 24, and ChIP-seq 10,
with different proportions of QC, peak calling, annotation, chromatin-state,
and downstream-analysis tasks. Because assay type, source workflow, and task
mix are coupled in this benchmark, we treat assay-level pass rates as
descriptive summaries rather than controlled estimates of intrinsic assay
difficulty.

\EndBody

\begin{center}
  \includegraphics[width=0.34\textwidth]{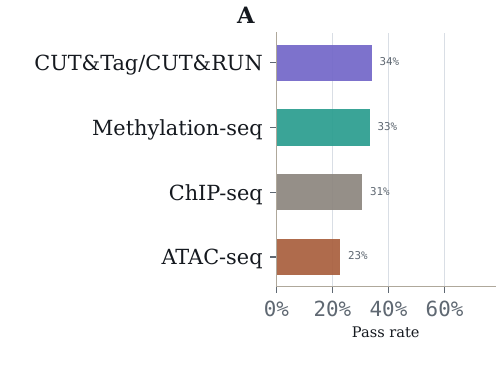}\hfill
  \includegraphics[width=0.62\textwidth]{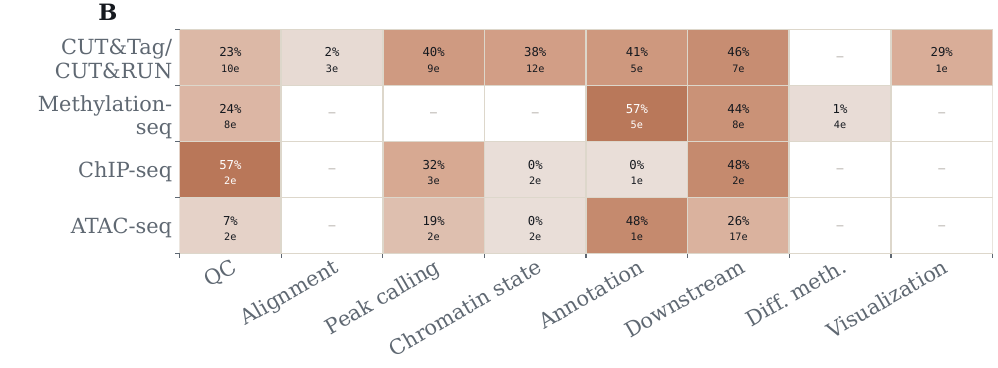}\\[0.3em]
  \includegraphics[width=0.62\textwidth]{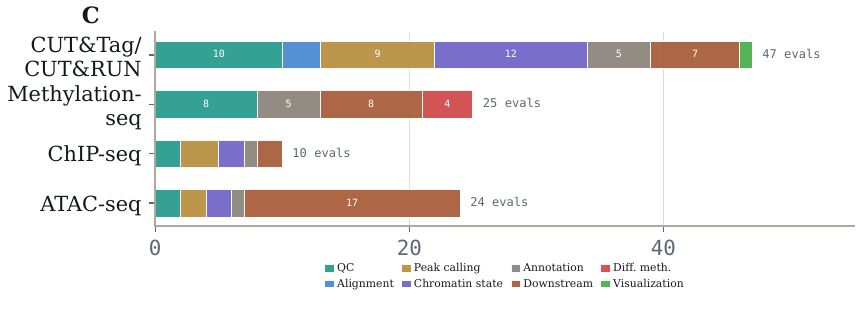}

  \captionof{figure}{\textbf{Performance varies by assay type.}
  Panel A: pass rate by assay type. Panel B:
  assay-by-task-category pass-rate
  heatmap; cell text reports pass rate and the number of evaluations in that
  cell. Panel C: evaluation-count composition by assay type, showing the task
  mix behind each assay summary.}
  \label{fig:assay-summary}
\end{center}

\StartBody

\subsection{Many failures trace to plausible but wrong scientific choices}

Endpoint failures clustered around concrete choices where a generic
bioinformatics workflow diverged from the assay-specific evidence in the
provided files. Manual review of 25 evaluations identified recurring failure
behaviors: using an incorrect statistic, applying a threshold incorrectly,
using the wrong unit, handling features or data layers incorrectly, mishandling
read or peak representation, and relying on a literature prior when
file-derived results supported a different answer
(Figure~\ref{fig:decision-traps}). These mistakes were often small in
implementation terms, but they determined whether the final structured answer
matched the data.

Component-level grading supported the same interpretation. Among 26,574 scored
answer fields from 5,051 trajectories, 68.2\% passed (18,124/26,574 fields),
compared with a 31.0\% endpoint pass rate (1,578/5,088 attempts). The gap indicates that
many failed endpoints contained some correct pieces, but used the wrong data
layer, statistic, unit, or source of evidence.

Examples span the assay families. In CUT\&RUN spike-in normalization, agents
had to decide how to align spike-in reads. End-to-end rather than local
Bowtie2 alignment \cite{langmead2012bowtie2} reduced spike-in recovery and
distorted downstream normalization. In WGBS outputs, agents had to decide how
to count each CpG: treating the two Bismark rows for a CpG dinucleotide
as independent sites \cite{krueger2011bismark} changed coverage and
differential-methylation statistics. In ATAC-seq, the choice was what to give
MACS3: using paired-end BAMs directly with \texttt{-f BAMPE} bypassed the
intended single-base Tn5 insertion-site representation
\cite{buenrostro2013atacseq,zhang2008macs}, shifting peak calls and downstream
motif or integration results. In interpretation tasks, agents sometimes used a
literature prior incorrectly, substituting a familiar mechanism for the
comparison supported by the files.

A related pattern appeared when agents computed the correct result but did not
submit it. Across reviewed GPT-5.5 trajectories for CUT\&Tag/CUT\&RUN and
methylation-seq, this occurred in seven evaluations; in four of those, the
correct answer was visible in the agent's own output before being replaced by a
more familiar workflow default or biological expectation. These cases separate
tool execution from scientific judgment: the result was available, but the
agent selected the less supported answer.

\EndBody

\begin{center}
  \includegraphics[width=0.74\textwidth]{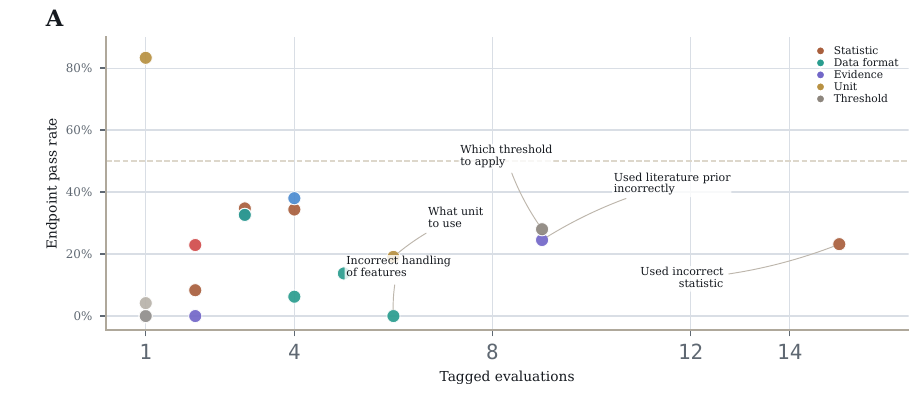}

  \vspace{0.18em}

  \includegraphics[width=0.84\textwidth]{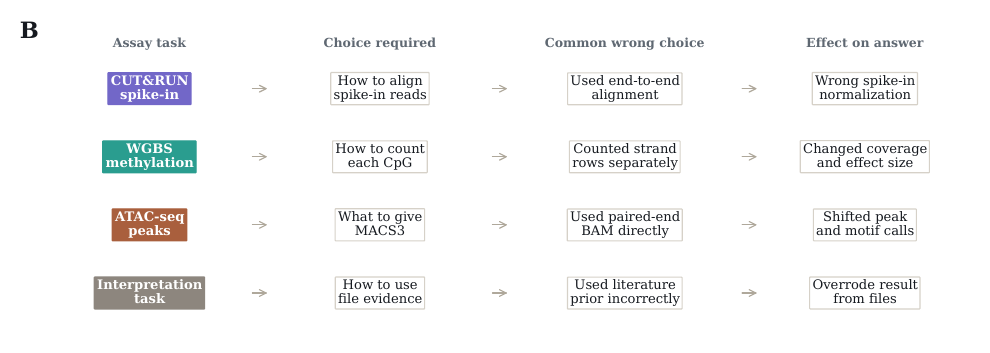}

  \captionof{figure}{\textbf{Many failures trace to plausible but wrong
  scientific choices.} Panel A plots manually tagged failure behaviors by the
  number of reviewed evaluations carrying each behavior and the corresponding
  endpoint pass rate. Panel B shows representative assay choices where a
  plausible default changed the endpoint answer. Manual review covers 25/106
  evaluations and is diagnostic rather than exhaustive.}
  \label{fig:decision-traps}
\end{center}

\StartBody

\section{Discussion}

EpiBench measures practical analysis work: can an agent recover a specific
empirical result from epigenomics data in the same way an experienced human
would? The results show that these local scientific decisions remain
unreliable. The strongest model-harness pair passed 45.0\% of endpoint attempts
(143/318 attempts; 95\% CI, 36.3--53.7), and many failures occurred despite
partial progress. Field-level scores and trajectory review suggest that agents
often found relevant files or computed useful intermediate values, but then
selected a familiar workflow default, literature-derived mechanism, or nearby
statistic that was not supported by the provided evidence. This separates
execution from judgment: an agent can operate the tools of an epigenomics
workflow and still submit the wrong biological answer.

The benchmark has several limitations. The task inventory is not balanced
across assay families: CUT\&Tag/CUT\&RUN (47 evaluations) and methylation-seq
(25) contribute more evaluations than ATAC-seq (24) or ChIP-seq (10), and
downstream analysis plus QC tasks dominate the benchmark. Some failure
behaviors also recur across related evaluations, so aggregate scores partially
reflect repeated tests of the same underlying decisions. This mirrors real
workflow failure modes, but it should be considered in evaluating benchmark
results.

Finally, deterministic graders intentionally constrain the answer surface.
They are useful for measurement, but they do not capture every scientifically
valid analysis path. EpiBench therefore should not be read as a complete test
of epigenomics reasoning. Rather, it defines a measurable sample of practical
epigenomics skills where current agents still fail in interpretable ways.
Future progress will require not only better tool use, but better grounding of
biological claims in the specific assay artifacts that support them.

\section{Methods}

\subsection{Evaluation construction}

Candidate evaluations were derived from real epigenomics analysis workflows.
An evaluation was retained when the target result could be reproduced from the
provided data, expressed through a constrained final answer, and graded
deterministically. Tasks were revised or removed when prompts over-specified
the method, when reasonable analysis choices produced incompatible answers, or
when the grader could not distinguish a supported result from plausible but
unsupported biological output.

\subsection{Agent runs and aggregation}

Each model-harness pair was run three times per evaluation. All 5,088
trajectory outputs were downloaded and parsed successfully. The result set
contains three attempts for every combination of 106 evaluations and 16
model-harness pairs. Nonpassing endpoint outputs are counted as nonpassing
attempts; no output was missing or invalid in this analysis. Component field scores were
available for 5,051 trajectories and are analyzed only as diagnostics.

\subsection{Statistical analysis}

The primary metric is evaluation-weighted endpoint pass rate. For each
model-harness pair and evaluation, we first compute the mean pass rate across
the three attempts, then report the mean across evaluations. Because each
model-harness pair has the same number of attempts for every evaluation, this
estimate is numerically identical to the raw run-level pass rate. Confidence
intervals use the Student $t$ distribution
over evaluation-level mean pass rates, matching the SpatialBench aggregation
convention \cite{workman2025spatialbench}. We report raw pass counts for
interpretability. Confidence intervals are computed over evaluations rather
than individual trajectories, because repeated attempts on the same evaluation
are not independent samples of the benchmark. We also report evaluation-level
attempt summaries: whether a model-harness pair passed any attempt, a
majority of attempts, or all three attempts for an evaluation. Component field
scores and failure-mode labels are reported as diagnostics rather than
benchmark scores.

\section*{Data availability}

Results files, public example evaluations, and representative trajectories
will be made available at
\url{https://github.com/latchbio/epibench}.

{\fontsize{8.3}{10.7}\selectfont

}

\EndBody

\end{document}